\documentclass[11pt,a4paper]{article}
\usepackage[hyperref]{acl2019}
\aclfinaltrue
\pdfoutput=1
\usepackage{times}
\usepackage{float}
\usepackage{latexsym}
\usepackage{xcolor,graphicx,soul,colortbl}
\RequirePackage[normalem]{ulem}
\RequirePackage{amsmath,amssymb,amsfonts,mathtools}
\RequirePackage[textsize=scriptsize,disable]{todonotes}
\RequirePackage{enumitem} \setlist{nosep}
\setlist[description]{font={\bfseries\slshape\color{gray!90!red}}}
\RequirePackage{tabularx}
\RequirePackage{tikz}
\usetikzlibrary{shapes,arrows,positioning,matrix,backgrounds,fit}
\RequirePackage{pgfplots}
\RequirePackage{pgfplotstable,subcaption}
\pgfplotsset{compat=1.14}
\pgfplotsset{nice/.style={grid=major, major grid style={line width=.2pt,draw=gray!30},}}
\RequirePackage[noend]{algpseudocode}
\RequirePackage[most]{tcolorbox}
\newtcolorbox{insight}[1][default]{boxsep=.5mm, left=0mm, right=0mm, top=0mm, bottom=0mm, colback=gray!4, colframe=gray!25, boxrule=.2mm, }
\intextsep \textfloatsep
\RequirePackage[skip=1ex plus.4ex minus.3ex]{caption}
\RequirePackage{titlesec}
\makeatletter
\g@addto@macro{\normalsize}{%
    \setlength{\abovedisplayskip}{1pt plus1pt}
    \setlength{\abovedisplayshortskip}{1pt plus1pt}
    \setlength{\belowdisplayskip}{1pt plus1pt}
    \setlength{\belowdisplayshortskip}{1pt plus1pt}}
\makeatother

\newcounter{magicrownumbers}
\newcommand\rownumber{\stepcounter{magicrownumbers}\arabic{magicrownumbers}}
\newcounter{mrownumbers}
\newcommand\rnumber{\stepcounter{mrownumbers}\arabic{mrownumbers}}

\RequirePackage{array}
\newcolumntype{L}[1]{>{\raggedright\let\newline\\\arraybackslash\hspace{0pt}}m{#1}}
\newcolumntype{C}[1]{>{\centering\let\newline\\\arraybackslash\hspace{0pt}}m{#1}}
\newcolumntype{R}[1]{>{\raggedleft\let\newline\\\arraybackslash\hspace{0pt}}m{#1}}

\def\ignore#1{}

\DeclareMathOperator*{\stability}{stability}
\DeclareMathOperator*{\wscore}{\mathit{R}}  
\DeclareMathOperator*{\sscore}{\mathit{Q}}  

\newcommand{\vek}[1]{{\bf {#1}}}

\newcommand{\DT}{{{\mathcal D_T}}}
\newcommand{\DS}{{{\mathcal D_S}}}
\newcommand{\ES}{{{\mathcal E}}}
\def\R{\mathbb{R}}

\newcommand{\tgt}{{{Tgt}}}
\newcommand{\src}{{{Src}}}
\newcommand{\srcinit}{{{SrcTune}}}
\newcommand{\concat}{{{Concat}}}
\newcommand{\avg}{{{Avg}}}

\newcommand{\yangC}{{{RegFreq}}}

\newcommand{\yang}{{RegFreqS}}
\newcommand{\ourReg}{{{RegSense}}}

\newcommand{\srctgt}{{{Src+Tgt}}}
\newcommand{\ourR}{{{\shortname:R}}}
\newcommand{\ourC}{{{\shortname:c}}}
\newcommand{\our}{{{\shortname}}}

\begin{document}
\colorlet{usercolorname}{green!10}
\sethlcolor{usercolorname}
\def\shortname{SrcSel}


\def\papertitle{Topic Sensitive Attention on Generic Corpora Corrects Sense Bias in Pretrained Embeddings}

\title{\papertitle}

\author{Vihari Piratla\thanks{\quad\texttt{vihari@cse.iitb.ac.in}} \\ IIT~Bombay \\\And
  Sunita Sarawagi \\ IIT~Bombay \\\And
  Soumen Chakrabarti \\ IIT~Bombay
}
\date{}
\maketitle
\begin{abstract}
Given a small corpus $\DT$ pertaining to a limited set of focused topics, our goal is to train embeddings that accurately capture the sense of words in the topic in spite of the limited size of $\DT$.  These embeddings may be used in various tasks involving~$\DT$.  
A popular strategy in limited data settings is to adapt pretrained embeddings $\ES$ trained on a large corpus. To correct for sense drift, fine-tuning, regularization, projection, and pivoting have been proposed recently.
Among these, regularization informed by a word's corpus frequency performed well, but we improve upon it using a new regularizer based on the stability of its cooccurrence with other words.
However, a thorough comparison across ten topics, spanning three tasks, with standardized settings of hyper-parameters, reveals that even the best embedding adaptation strategies provide small gains beyond well-tuned baselines, which many earlier comparisons ignored.
In a bold departure from adapting pretrained embeddings, we propose using $\DT$ to probe, attend to, and borrow fragments from any large, topic-rich source corpus (such as Wikipedia), which need not be the corpus used to pretrain embeddings.  This step is made scalable and practical by suitable indexing.
We reach the surprising conclusion that even limited corpus augmentation is more useful than adapting embeddings, which suggests that non-dominant sense information may be irrevocably obliterated from pretrained embeddings and cannot be salvaged by adaptation.
%
%
All our code and data splits will be made publicly available at \url{https://github.com/vihari/focussed_embs}.
\end{abstract}


\section{Introduction}
\label{sec:Intro}

Word embeddings \cite{MikolovSCCD2013word2vec,PenningtonSM2014GloVe} benefit many natural language processing (NLP) tasks.
Often, a group of tasks may involve a limited corpus $\DT$ pertaining to a few focused topics, e.g., discussion boards on Physics, video games, or Unix, or a forum for discussing medical literature.
Because $\DT$ may be too small to train word embeddings to sufficient quality, a prevalent practice is to harness general-purpose embeddings $\ES$ pretrained on a broad-coverage corpus, not tailored to the topics of interest.
The pretrained embeddings are sometimes used as-is (`pinned').  Even if $\ES$ is trained on a `universal' corpus, considerable sense shift may exist in the meaning of polysemous words and their cooccurrences and similarities with other words.  In a corpus about Unix, `cat' and `print' are more similar than in Wikipedia.  `Charge' and `potential' are more related in a Physics corpus than in Wikipedia.  Thus, pinning can lead to poor target task performance in case of serious sense mismatch.  
Another popular practice is to initialize the target embeddings to the pretrained vectors, but then ``fine-tune'' using $\DT$ to improve performance in the target~\cite{MouPLXZJ15,min17,Howard2018}.
As we shall see, the number of epochs of fine-tuning is a sensitive knob --- excessive fine-tuning might lead to ``catastrophic forgetting'' \citep{kirkpatrick2017overcoming} of useful word similarities in ~$\ES$, and too little fine-tuning may not adapt to target sense.

Even if we are given development (`dev') sets for target tasks, the best balancing act between a pretrained $\ES$ and a topic-focused $\DT$ is far from clear.
Should we fine-tune (all word vectors) in epochs and stop when dev performance deteriorates?  
Or should we keep some words close to their pretrained embeddings (a form of regularization) and allow others to tune more aggressively?  On what properties of $\ES$ and $\DT$ should the regularization strength of each word depend?  Our first contribution is a new measure of semantic drift of a word from $\ES$ to $\DT$, which can be used to control the regularization strength.  In terms of perplexity, we show that this is superior to both epoch-based tuning, as well as regularization based on simple corpus frequencies of words \citep{regemnlp}.  
Yet another option is to learn projections to align generic embeddings to the target sense~\cite{BollegalaMK15,Barnes2018,Sarma2018}, or to a shared common space~\cite{YinS16,CoatesB18,BaoB2018}
However, in carefully controlled experiments, none of the proposed approaches to adapting pretrained embeddings consistently beats the trivial baseline of discarding them and training afresh on~$\DT$!


Our second contribution is to explore other techniques beyond adapting generic embeddings $\ES$.  Often, we might additionally have easy access to a broad corpus $\DS$ like Wikipedia.  $\DS$ may span many diverse topics, while $\DT$ focuses on one or few, so there may be  large \emph{overall} drift from $\DS$ to $\DT$ too.  However, a judicious \emph{subset} $\widehat{\DS} \subset \DS$ may exist that would be excellent for augmenting~$\DT$.   The large size of $\DS$ is not a problem: we use an inverted index that we probe with documents from $\DT$ to efficiently identify~$\widehat{\DS}$.  Then we apply a novel perplexity-based joint loss over $\widehat{\DS}\cup\DT$ to fit adapted word embeddings. 
While most of recent research focus has been on designing better methods of adapting pretrained embeddings, we show that retraining with selected source \emph{text} is significantly more accurate than the best of embeddings-only strategy, while runtime overheads are within practical limits.

An important lesson is that non-dominant sense information may be irrevocably obliterated from generic embeddings; it may not be possible to salvage this information by post-facto adaptation.

   

\noindent
Summarizing, our contributions are:
\begin{itemize}[partopsep=0ex,topsep=0ex,leftmargin=*]
\item We propose new formulations for training topic-specific embeddings on a limited target corpus $\DT$ by (1) adapting generic pre-trained word embeddings $\ES$, and/or (2) selecting from any available broad-coverage corpus $\DS$.
\item We perform a systematic comparison of our and several recent methods on three tasks spanning ten topics and offer many insights.

\item Our selection of $\widehat{\DS}$ from $\DS$ and joint perplexity minimization on $\widehat{\DS} \cup \DT$ perform better than pure embedding adaptation methods, at the (practical) cost of processing~$\DS$.
\item We evaluate our method even with contextual embeddings.  The relative performance of the adaptation alternatives remain fairly stable whether the adapted embeddings are used on their own, or concatenated with context-sensitive embeddings~\citep{PetersNIGCKZ2018ELMo, Cer+2018UnivSentEncoder}.
\end{itemize}

\section{Related work and baselines}
\label{sec:Related}

\subsubsection*{CBOW}

We review the popular CBOW model for learning unsupervised word representations~\citep{MikolovSCCD2013word2vec}.  As we scan the corpus, we collect a \emph{focus} word $w$ and a set $C$ of \emph{context} words around it, with corresponding embedding vectors $\pmb{u}_w \in \R^n$ and $\pmb{v}_c \in \R^n$, where $c \in C$.  The two embedding matrices $\pmb{U}, \pmb{V}$ are  estimated as:
\begin{align}
\max_{\pmb{U}, \pmb{V}} \hspace{-.7em}
\sum_{\langle w, C\rangle \in \mathcal{D}} \hspace{-.7em}
\sigma(\pmb{u}_w \cdot \pmb{v}_C)
+ \sum_{\bar w \sim \mathcal{D}} \sigma(-\pmb{u}_{\bar{w}} \cdot \pmb{v}_C)
\label{eq:cbow}
\end{align}
Here $\pmb{v}_C$ is the average of the context vectors in~$C$. $\bar w$ is a negative focus word sampled from a slightly distorted  unigram distribution of~$\mathcal{D}$.  Usually downstream applications use only the embedding matrix $\pmb{U}$, with each word vector scaled to unit length. Apart from CBOW, \citet{MikolovSCCD2013word2vec} defined the related skipgram model, and \cite{PenningtonSM2014GloVe} proposed the Glove model, which can also be used in our framework.  We found CBOW to work better for our downstream tasks.

\subsubsection*{\src, \tgt{} and \concat{} baselines}

In the `\src' option, pre-trained embeddings $\pmb{u}^S_w$ trained only on a large corpus are used as-is.  The other extreme, called `\tgt', is to train word embeddings from scratch on the limited target corpus~$\DT$. In our experiments we found that \src\ performs much worse than \tgt, indicating the presence of significant drift in prominent word senses.   Two other simple baselines, are `\concat', that concatenates the source and target trained embeddings and let the downstream task figure out their relative roles, and '\avg' that following \cite{CoatesB18} takes their simple average.  Another option is to let the downstream task learn to combine multiple embeddings as in \cite{ZhangRW2016}.

As word embeddings have gained popularity for representing text in learning models, several methods have been proposed for enriching small datasets with pre-trained embeddings.

\subsubsection*{Adapting pre-trained embeddings}

\paragraph{\srcinit:}
A popular method~\citep{min17, WangHF17,Howard2018} is to use the source embeddings $\pmb{u}^S_w$ to initialize $\pmb{u}_w$ and thereafter train on~$\DT$.  We call this `\srcinit'.  Fine-tuning requires careful control of the number of epochs with which we train on~$\DT$.  Excessive training can wipe out any benefit of the source because of catastrophic forgetting.  Insufficient training may not incorporate target corpus senses in case of polysemous words, and adversely affect target tasks~\citep{MouPLXZJ15}. The number of epochs can be controlled using perplexity on a held-out $\DT$, or using downstream tasks.   \citet{Howard2018} propose to fine-tune a whole language model using careful differential learning rates.  However, epoch-based termination may be inadequate.  Different words may need diverse trade-offs between the source and target topics, which we discuss next.


\paragraph{\yangC~(frequency-based regularization):}
\citet{regemnlp} proposed to train word embeddings using $\DT$, but with a regularizer to prevent a word $w$'s embedding from drifting too far from the source embedding ($\pmb{u}^S_w$).  The weight of the regularizer is meant to be inversely proportional to the concept drift of $w$ across the two corpus.  Their limitation was that corpus frequency was used as a surrogate for stability; high stability was awarded to only words frequent in both corpora.  As a consequence, very few words in a focused $\DT$ about Physics will benefit from a broad coverage corpus like Wikipedia.  Thousands of words like {\em galactic, stars, motion, x-ray,} and {\em momentum} will get low stability, although their prominent sense is the same in the two corpora.  We propose a better regularization scheme in this paper.  Unlike us, \citet{regemnlp} did not compare with fine-tuning.

\paragraph{Projection-based methods} attempt to project embeddings of one kind to another, or to a shared common space.  \citet{Bollegala2014} and \citet{Barnes2018} proposed to learn a linear transformation between the source and target embeddings.
\citet{YinS16} transform multiple embeddings to a common `meta-embedding' space. Simple averaging are also shown to be effective \citep{CoatesB18}, and a recent~\cite{BaoB2018} auto-encoder based meta-embedder (AEME) is the state of the art. \citet{Sarma2018} proposed CCA to project both embeddings to a common sub-space. Some of these methods designate a subset of the overlapping words as pivots to bridge the target and source parameters in various ways~\cite{BlitzerMP06, Ziser2018, BollegalaMK15}. Many such techniques were proposed in a cross-domain setting, and specifically for the sentiment classification task. Gains are mainly from effective transfer of sentiment representation across domains.  Our challenge arises when a corpus with broad topic coverage pretrains dominant word senses quite different from those needed by tasks associated with narrower topics.






\subsubsection*{Language models for task transfer}
Complementary to the technique of adapting individual word embeddings is the design of deeper sequence models for task-to-task transfer.  \citet{Cer+2018UnivSentEncoder,SubramanianTBP2018SentenceMTL} propose multi-granular transfer of sentence and word representations across tasks using  Universal Sentence Encoders. 
ELMo \citep{PetersNIGCKZ2018ELMo} trains a multi-layer sequence model to build  a context-sensitive representation of words in a sentence. 
ULMFiT \citep{Howard2018} present additional tricks such as gradual unfreezing of parameters layer-by-layer, and exponentially more aggressive fine-tuning toward output layers. 
\citet{devlin2018bert} propose a deep bidirectional language model for generic contextual word embeddings.  We show that our topic-sensitive embeddings provide additional benefit even when used with contextual embeddings.

\section{Proposed approaches}
\label{sec:Approach}

We explore two families of methods: (1)~those that have access to only pretrained embeddings (Sec~\ref{sec:Approach:StabilityReg}), and (2)~those that also have access to a source corpus with broad topic coverage (Sec~\ref{sec:Approach:WithSource}). 


\subsection{\ourReg: Stability-based regularization}
\label{sec:Approach:StabilityReg}

Our first contribution is a more robust definition of stability to replace the frequency-based regularizer of \yangC.
We first train word vectors on $\DT$, and assume the pretrained embeddings $\ES$ are available.
Let the focus embeddings of word $w$ in $\ES$ and $\DT$ be 
$\pmb{u}^S_w$ and $\pmb{u}^T_w$.
We overload $\ES \cap \DT$ as words that occur in both.
For each word $w \in \ES \cap \DT$, we
compute $N^{(K)}_S(w, \ES\cap\DT)$, the $K$ nearest neighbors of $w$ with respect to the generic embeddings, i.e.,
with the largest values of $\cos(\pmb{u}^S_w, \pmb{u}^S_n)$ from $\ES\cap\DT$.
Here $K$ is a suitable hyperparameter.
Now we define $\stability(w)=$
\begin{align}
\frac{\sum_{n\in N^{(K)}_S(w,\ES\cap\DT)} \cos(\pmb{u}_w^T, \pmb{u}_n^T)}
{|N^{(K)}_S(w, \ES \cap \DT)|}
\label{eqn:sim_score}
\end{align}
Intuitively, if we consider near neighbors $n$ of $w$ in terms of source embeddings, and most of these $n$'s have target embeddings very similar to the target embedding of $w$, then $w$ is stable across $\ES$ and $\DT$, i.e., has low semantic drift from $\ES$ to~$\DT$.

While many other forms of $\stability$ can achieve the same ends, ours seems to be the first formulation that goes beyond mere word frequency and employs the topological stability of near-neighbors in the embedding space.  Here is why this is important.  Going from a  generic corpus like Wikipedia to the very topic-focused StackExchange (Physics) corpus $\DT$, the words {\em x-ray, universe, kilometers, nucleons, absorbs, emits, sqrt, anode, diodes}, and {\em km/h} have large stability per our definition above, but low stability according to \citeauthor{regemnlp}'s frequency method since they are (relatively) rare in source.  Using their method, therefore, these words will not benefit from reliable pretrained embeddings.


Finally, the word regularization weight is:
\begin{align}
\wscore(w) &= \max(0, \tanh\bigl(\lambda \, \stability(w))\bigr).
\label{eqn:sem_drift}
\end{align}
Here $\lambda$ is a hyperparameter.
$\wscore(w)$ above is a replacement for the regularizer used by \citet{regemnlp}.  
If $\wscore(w)$ is large, it is regularized more heavily toward its source embedding, keeping $\pmb{u}_w$ closer to~$\pmb{u}^S_w$.  The modified CBOW loss is:
\begin{multline}
\max_{\pmb{U}, \pmb{V}} \hspace{-.7em}
\sum_{\langle w, C\rangle \in \mathcal{D}} \hspace{-.7em}
\sigma(\pmb{u}_w \cdot \pmb{v}_C) 
+ \sum_{\bar w \sim \mathcal{D}} \sigma(-\pmb{u}_{\bar{w}} \cdot \pmb{v}_C) \\ + \sum_w
\colorbox{green!8}{$\wscore(w)$}
\|\pmb{u}_w - \pmb{u}^S_w \|^2
\label{eq:RegSense}
\end{multline}
Our $\wscore(w)$ performs better than \citeauthor{regemnlp}'s.


\subsection{Source selection and joint perplexity}
\label{sec:Approach:WithSource}

To appreciate the limitations of regularization, consider words like {\em potential, charge, law, field, matter, medium,} etc.  These will get small stability ($\wscore(w)$) values because their dominant senses in a universal corpus do not match with those in a Physics corpus~($\DT$), but $\DT$ may be too limited to wipe that dominant sense for a subset of words while preserving the meaning of stable words.
However, there are plenty of high-quality broad-coverage sources like Wikipedia that includes plenty of Physics documents that could gainfully supplement~$\DT$.  Therefore, we seek to include target-relevant documents from a generic source corpus $\DS$, even if the dominant sense of a word in $\DS$ does not match that in~$\DT$.  The goal is to do this without solving the harder problem of unsupervised, expensive and imperfect sense discovery in $\DS$ and sense tagging of~$\DT$, and using per-sense embeddings.

The main steps of the proposed approach, \shortname, are shown in 
\figurename~\ref{fig:SrcSel}.  Before describing the steps in detail, we note that preparing and probing a standard inverted index \citep{BaezaYatesR1999MIR} are extremely fast, owing to decades of performance optimization.  Also, index preparation can be amortized over multiple target tasks.  (The granularity of a `document' can be adjusted to the application.)

\begin{figure}[th]
\centering
\begin{tcolorbox}[boxsep=0mm,colback=gray!4,colframe=gray!25,boxrule=.2mm]
\begin{algorithmic}[1] \raggedright
\State Index all source docs $\DS$ in a text retrieval engine.
\State Initialize a score accumulator $a_{s}$ for each source doc $s\in\DS$.
\For{each target doc $t \in \DT$}
\State Get source docs most similar to~$t$.
\State Augment their score accumulators.
\EndFor
\State $\widehat{\DS} \leftarrow \varnothing$
\For{each source doc $s \in \DS$}
\If{$a_{s}$ is ``sufficiently large''}
\State Add $s$ to $\widehat{\DS}$.
\EndIf
\EndFor
\State Fit word embeddings to optimize a joint objective over $\widehat{\DS}\cup\DT$.
\end{algorithmic}
\end{tcolorbox}
\caption{Main steps of \shortname.}
\label{fig:SrcSel}
\end{figure}

\paragraph{Selecting source documents to retain:}
Let $s \in \DS, t \in \DT$ be source and target documents.  Let $\text{sim}(s,t)$ be the similarity between them, in terms of the TFIDF cosine score commonly used in Information Retrieval \citep{BaezaYatesR1999MIR}.  The total vote of $\DT$ for $s$ is then $\sum_{t \in \DT}\text{sim}(s, t)$.  We choose a suitable cutoff on this aggregate score, to reduce $\DS$ to $\widehat{\DS}$, as follows.  Intuitively, if we hold out a randomly sampled part of $\DT$, our cutoff should let through a large fraction (we used 90\%) of the held-out part.  Once we find such a cutoff, we apply it to $\DS$ and retain the source documents whose aggregate scores exceed the cutoff.  Beyond mere selection, we design a joint perplexity objective over $\widehat{\DS}\cup\DT$, with a term for the amount of trust we place in a retained source document. This limits damage from less relevant source documents that slipped through the text retrieval filter. Since the retained documents are weighted based on their relevance to the topical target corpus $\DT$, we found it beneficial to also include a percentage (we used 10\%) of randomly selected documents from $\DS$. We refer to the method that only uses documents retained using text retrieval filter as \ourR{} and only randomly selected documents from $\DS$ as \ourC. \our{} uses documents both from the retrieval filter and random selection. 

\paragraph{Joint perplexity objective:}
Similar to Eqn.~\eqref{eq:cbow}, we will sample word and context $\langle w, C\rangle$ from $\DT$ and $\widehat{\DS}$.  Given our limited trust in $\widehat{\DS}$, we will give each sample from $\widehat{\DS}$ an alignment score $\sscore(w,C)$.  This should be large when $w$ is used in a context similar to contexts in~$\DT$.  We judge this based on the target embedding~$\pmb{u}^T_w$:
\begin{align}
\sscore(w, C) &=
\max\left\{0, \cos\left(\pmb{u}^T_w, \pmb{v}^T_C \right) \right\}.
\label{eq:cscore}
\end{align}
Since $\pmb{u}_w$ represents the sense of the word in the target, source contexts $C$ which are similar will get a high score.  Similarity in source embeddings is not used here because our intent is to preserve the target senses.  We tried other forms such as dot-product or its exponential and chose the above form because it is bounded and hence less sensitive to gross noise in inputs. 

The word2vec objective~\eqref{eq:cbow} is enhanced to
\begin{multline}
\sum_{\langle w, C\rangle \in \DT}
\hspace{-1em} \left[
\sigma(\pmb{u}_w \cdot \pmb{v}_C) + \textstyle
\sum_{\bar{w} \sim \DT} \sigma(-\pmb{u}_{\bar{w}} \cdot \pmb{v}_C)
\right] \\[-1ex]
+ \sum_{\langle w, C\rangle \in \widehat{\DS}} \!\!\!\!
\colorbox{red!7}{$\sscore(w,C)$} \, \Bigl[
\sigma(\pmb{u}_w \cdot \pmb{v}_C) +  \\[-2ex]
\textstyle \sum_{\bar{w} \sim \widehat{\DS}} \! 
\sigma(-\pmb{u}_{\bar{w}} \cdot \pmb{v}_C) \Bigr]. \label{eq:SnippetSelect}
\end{multline}
The first sum is the regular word2vec loss over~$\DT$.  Word $\bar{w}$ is sampled from the vocabulary of $\DT$ as usual, according to a suitable distribution.  The second sum is over the retained source documents~$\widehat{\DS}$. Note that $\sscore(w,C)$ is computed using the pre-trained target embeddings and does not change during the course of training.


\paragraph{\shortname+\ourReg{} combo:}
Here we combine objective \eqref{eq:SnippetSelect} with the regularization term in \eqref{eq:RegSense}, where $\wscore$ uses all of~$\ES$ as in \ourReg.





\section{Experiments}
\label{sec:Expt}

We compare the methods discussed thus far, with the goal of answering these research questions:
\begin{enumerate}[partopsep=0ex,topsep=0ex,leftmargin=*]
\item Can word-based regularization (\yangC\ and \ourReg) beat careful termination at epoch granularity, after initializing with source embeddings (\srcinit)?
\item How do these compare with just fusing \src\ and \tgt\ via recent meta-embedding methods like AAEME~\cite{BaoB2018}\footnote{We used the implementation available at: https://github.com/CongBao/AutoencodedMetaEmbedding}?
\item Does \our\ provide sufficient and consistent gains over \ourReg\ to justify the extra effort of processing a  source corpus?
\item Do contextual embeddings obviate the need for adapting word embeddings?
\end{enumerate}
We also establish that initializing with source embeddings also improves regularization methods. (Curiously, \yangC\ was never combined with source initialization.)

\subsubsection*{Topics and tasks}
\label{sec:DedupDataset}
We compare across 15 topic-task pairs spanning 10 topics and 3 task types: an unsupervised language modeling task on five topics, a document classification task on six topics, and a duplicate question detection task on four topics. 
In our setting, $\DT$ covers a small subset of topics in $\DS$, which is the 20160901\footnote{The target corpora in our experiments came from datasets that were created before this time.} version dump of Wikipedia.  Our tasks are different from GLUE-like multi-task learning \citep{WangSMHLB2019glue}, because our focus is on the problems created by the divergence between prominent sense-dominated generic word embeddings and their sense in narrow target topics.  We do not experiment on the cross-domain sentiment classification task popular in domain adaptation papers since they benefit more from sharing sentiment-bearing words, than learning the correct sense of polysemous words, which is our focus here.  All our experiments are on public datasets, and we will publicly release our experiment scripts and code. 


\paragraph{StackExchange topics}

We pick four topics (Physics, Gaming, Android and Unix) from the CQADupStack\footnote{\protect\url{http://nlp.cis.unimelb.edu.au/resources/cqadupstack/}} dataset of questions and responses.  For each topic, the available response text is divided into $\DT$, used for training/adapting embeddings, and $\widetilde{\DT}$, the evaluation fold used to measure perplexity.
In each topic, the target corpus $\DT$ has 2000 responses totalling roughly 1 MB. We also report results with changing sizes of $\DT$. Depending on the method we use $\DT, \DS$, or $\pmb{u}^S$ to train topic-specific embeddings and evaluate them as-is on two tasks that train task-specific layers on top of these fixed embeddings.
%
%
\noindent 
The first is an {\bf unsupervised language modeling task} where we train a LSTM\footnote{\url{https://github.com/tensorflow/models/blob/master/tutorials/rnn/ptb/ptb_word_lm.py}} on the adapted embeddings (which are pinned) and report perplexity on~$\widetilde{\DT}$.
The second is a {\bf Duplicate question detection task.}
Available in each topic are human annotated duplicate questions (statistics in \tablename~\ref{tab:stats} of Appendix) which we partition across 
train, test and dev as 50\%, 40\%, 10\%.  For contrastive training, we add four times as much randomly chosen non-duplicate pairs.  The goal is to predict duplicate/not for a question pair, for which we use word mover distance \citep[WMD]{KusnerSKW15} over adapted word embeddings.  We found WMD more accurate than BiMPM~\citep{WangHF17}.  
We use three splits of the target corpus, and for each resultant embedding, measure AUC on three random (train-)dev-test splits of question pairs, for a total of nine runs.  For reporting AUC, WMD does not need the train fold.

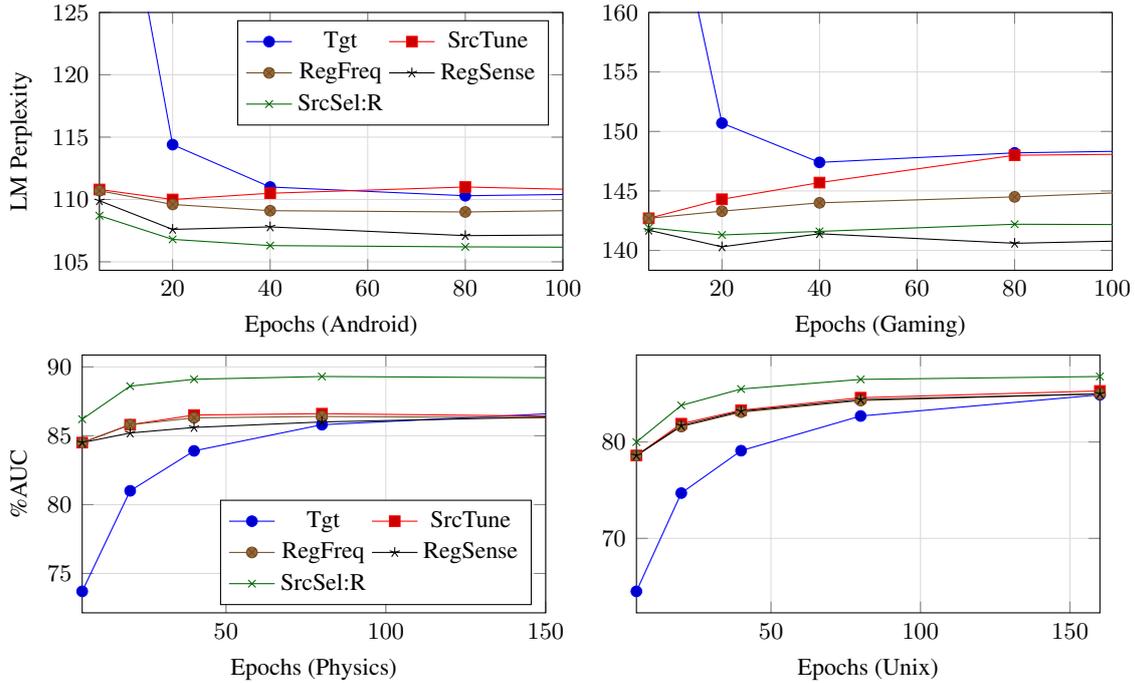
\begin{figure*}[t]
\pgfplotstableread{images/physics-ppl.dat} \physicsppl
\pgfplotstableread{images/unix-ppl.dat} \unixppl
\pgfplotstableread{images/android-ppl.dat} \androidppl
\pgfplotstableread{images/gaming-ppl.dat} \gamingppl
\pgfplotstableread{images/physics-auc.dat} \physicsauc
\pgfplotstableread{images/unix-auc.dat} \unixauc
\pgfplotstableread{images/android-auc.dat} \androidauc
\pgfplotstableread{images/gaming-auc.dat} \gamingauc
\def\chartheight{50mm}
\centering
\begin{subfigure}[b]{0.48\textwidth}
\begin{tikzpicture}[font=\small]
\begin{axis}[nice, width=\hsize, height=\chartheight, xlabel=Epochs (Android), ylabel=LM Perplexity, legend pos=north east, xmin=5, xmax=100,ymax=125, legend columns=2]
\addplot table [y=tgt, x=epoch] from \androidppl;
\addlegendentry{\tgt}
\addplot table [y=srcinit, x=epoch] from \androidppl;
\addlegendentry{\srcinit}
\addplot table [y=yangC, x=epoch] from \androidppl;
\addlegendentry{\yangC}
\addplot table [y=regour, x=epoch] from \androidppl;
\addlegendentry{\ourReg}
\addplot[color=green!40!black,mark=x] table [y=retr, x=epoch] from \androidppl;
\addlegendentry{\ourR}
\end{axis}
\end{tikzpicture}
\end{subfigure}
\begin{subfigure}[b]{0.48\textwidth}
\begin{tikzpicture}[font=\small]
\begin{axis}[nice, width=\hsize, height=\chartheight, xlabel=Epochs (Gaming), legend pos=north east, xmin=5, xmax=100, ymax=160]
\addplot table [y=tgt, x=epoch] from \gamingppl;
\addplot table [y=srcinit, x=epoch] from \gamingppl;
\addplot table [y=yangC, x=epoch] from \gamingppl;
\addplot table [y=regour, x=epoch] from \gamingppl;
\addplot[color=green!40!black,mark=x] table [y=retr, x=epoch] from \gamingppl;
\end{axis}
\end{tikzpicture}
\end{subfigure}
\begin{subfigure}[b]{0.48\textwidth}
\begin{tikzpicture}[font=\small]
\begin{axis}[nice, width=\hsize, height=\chartheight, xlabel=Epochs (Physics), ylabel={\%AUC}, legend pos=south east, xmin=5, xmax=150, legend columns=2]
\addplot table [y=tgt, x=epoch] from \physicsauc;
\addlegendentry{\tgt}
\addplot table [y=srcinit, x=epoch] from \physicsauc;
\addlegendentry{\srcinit}
\addplot table [y=yangC, x=epoch] from \physicsauc;
\addlegendentry{\yangC}
\addplot table [y=regour, x=epoch] from \physicsauc;
\addlegendentry{\ourReg}
\addplot[color=green!40!black,mark=x] table [y=retr, x=epoch] from \physicsauc;
\addlegendentry{\ourR}
\end{axis}
\end{tikzpicture}
\end{subfigure}
\begin{subfigure}[b]{0.48\textwidth}
\begin{tikzpicture}[font=\small]
\begin{axis}[nice, width=\hsize, height=\chartheight, xlabel=Epochs (Unix), legend pos=south east, xmin=5, xmax=160]
\addplot table [y=tgt, x=epoch] from \unixauc;
\addplot table [y=srcinit, x=epoch] from \unixauc;
\addplot table [y=regour, x=epoch] from \unixauc;
\addplot table [y=yangC, x=epoch] from \unixauc;
\addplot[color=green!40!black,mark=x] table [y=retr, x=epoch] from \unixauc;
\end{axis}
\end{tikzpicture}
\end{subfigure}
\caption{Language model perplexity (top row) and AUC on duplicate question detection (bottom row).}
\label{fig:auc}
\end{figure*}

\paragraph{Medical domain:}
This domain from the Ohsumed\footnote{\url{https://www.mat.unical.it/OlexSuite/Datasets/SampleDataSets-about.htm}} dataset has abstracts on cardiovascular diseases.  We sample 1.4\,MB of abstracts as target corpus~$\DT$.  We evaluate embeddings on two tasks: (1)~unsupervised language modeling on remaining abstracts, and (2)~supervised classification on 23 MeSH classes based on title.  We randomly select 10,000 titles with train, test, dev split as 50\%, 40\%, and 10\%.
Following~\citet{JoulinGBM17}, we train a softmax layer on the average of adapted (and pinned) word embeddings.


\paragraph{Topics from 20~newsgroup}
We choose the five top-level classes in the 20~newsgroup dataset\footnote{\url{http://qwone.com/~jason/20Newsgroups/}} as topics; viz.: \emph{Computer, Recreation, Science, Politics, Religion}.  The corresponding five downstream tasks are text classification over the 3--5 fine-grained classes under each top-level class.
Train, test, dev splits were 50\%, 40\%, 10\%.  We average over nine splits.  
The \texttt{body} text is used as $\DT$ and \texttt{subject} text is used for classification.

Pretrained embeddings $\ES$  are trained on  Wikipedia using the default settings of word2vec's CBOW model.

\begin{table}
\setlength\tabcolsep{1.0pt}
\centering
\begin{tabular}{|l|c|c|c|c|}
    \hline 
    Method & Physics & Gaming & Android & Unix \\
    \hline
    \tgt & 121.9 & 185.0 & 142.7 & 159.5 \\
    \tgt (unpinned) & -0.6 & -0.8 & 0.2 & 0.1 \\
    \hline
\end{tabular}
    \caption{Average reduction in perplexity, when embeddings are not pinned,
    on four Stackexchange topics.}
    \label{tab:lm_ft}
\end{table}

\subsection{Effect of fine-tuning embeddings on the target task}
We chose to pin embeddings in all our experiments, once adapted to the target corpus, namely the document classification task on medical and 20 newsgroup topics and language model task on five different topics. This is because we did not see any improvements when we unpin the input embeddings. We summarize in Table~\ref{tab:lm_ft} the results when the embeddings are not pinned on language model task on the four StackExchange topics.  

\subsection{Epochs vs.\ regularization results}

In Figure~\ref{fig:auc} we show perplexity and AUC against training epochs.  Here we focus on four methods: \tgt, \srcinit, \yangC,  and \ourReg.  First note that \tgt\ continues to improve on both perplexity and AUC metrics beyond five epochs (the default in word2vec code\footnote{\url{https://code.google.com/archive/p/word2vec/}} and left unchanged in \yangC{}\footnote{\url{https://github.com/Victor0118/cross_domain_embedding/}} \citep{regemnlp}).  In contrast, \srcinit, \ourReg, and \yangC\ are much better than \tgt\ at five epochs, saturating quickly.  With respect to perplexity, \srcinit\ starts getting worse around 20 iterations and becomes identical to \tgt, showing catastrophic forgetting.   Regularizers in \yangC\ and \ourReg\ are able to reduce such forgetting, with \ourReg\ being more effective than \yangC.
These experiments show that any comparison that chooses a fixed number of training epochs across all methods is likely to be unfair.  Henceforth we will use a validation set for the stopping criteria.  While this is standard practice for supervised tasks, most word embedding code we downloaded ran for a fixed number of epochs, making comparisons unreliable.  We conclude that validation-based stopping is critical for fair evaluation.

\begin{table}[ht]
\setlength\tabcolsep{1.0pt}
\begin{tabular}{|l|c|c|c|c|c|}
\hline 
Method & Physics & Gaming & Android & Unix & Med \\
\hline
\tgt & 121.9 & 185.0 & 142.7 & 159.5 & 158.9 \\
\hline
\srcinit & $2.3$&$6.8$&$1.1$&$3.1$&$5.5$\\
\yangC &$2.1$&$7.1$&$1.8$&$3.4$&$6.8$\\
\ourReg &$5.0$&$13.8$&$6.7$&$9.7$&$14.6$\\
\our &$5.8$&$11.7$&$5.9$&$6.4$&$8.6$\\
\hline
\our &$6.2$&$12.5$&$7.9$&$9.3$&$10.5$\\
+\ourReg & & & & & \\
\hline
\end{tabular}
\caption{\label{tab:perplexity}Average reduction in language model perplexity over \tgt\ on five topics. $\pm$ standard deviation are shown in \tablename~\ref{tab:perplexity_with_sd} in the Appendix}
\end{table}


\begin{table}[ht]
\setlength\tabcolsep{3.0pt}
\begin{tabular}{|l|l|l|l|l|}
\hline
 & Physics & Gaming & Android & Unix 
 \\
\hline
\tgt & 86.7 & 82.6 & 86.8 & 85.4 
\\
\hline
\src & -2.3$_{\pm \text{0.5}}$ & 0.8$_{\pm \text{0.5}}$ & -3.7$_{\pm \text{0.5}}$ & -7.1$_{\pm \text{0.3}}$ 
\\
\concat & -1.1$_{\pm \text{0.5}}$ & 1.4$_{\pm \text{0.3}}$ & -2.1$_{\pm \text{0.3}}$ & -4.5$_{\pm \text{0.4}}$ 
\\
AAEME & 1.2$_{\pm \text{0.2}}$ & 4.6$_{\pm \text{0.0}}$ & -0.3$_{\pm \text{0.2}}$ & 0.0$_{\pm \text{0.2}}$ 
\\
\srcinit & -0.3$_{\pm \text{0.3}}$ & 1.9$_{\pm \text{0.2}}$ & 0.6$_{\pm \text{0.2}}$ & -0.0$_{\pm \text{0.2}}$ 
\\
\yangC & -0.4$_{\pm \text{0.2}}$ & 2.4$_{\pm \text{0.2}}$ & -0.5$_{\pm \text{0.5}}$ & -0.5$_{\pm \text{0.2}}$ 
\\
\ourReg & -0.4$_{\pm \text{0.5}}$ & 2.2$_{\pm \text{0.1}}$ & -0.5$_{\pm \text{0.5}}$ & -0.5$_{\pm \text{0.4}}$ 
\\
\our & 3.6$_{\pm \text{0.2}}$ & 3.0$_{\pm \text{0.2}}$ & 0.8$_{\pm \text{0.3}}$ & 2.1$_{\pm \text{0.2}}$ \\
\rowcolor{green!10}
\our & 3.6$_{\pm \text{0.2}}$ & 3.1$_{\pm \text{0.5}}$ & 0.8$_{\pm \text{0.3}}$ &  2.1$_{\pm \text{0.2}}$ \\ 
\rowcolor{green!10} +\ourReg & & & & \\
\hline
\end{tabular}
\caption{\label{tab:auc}AUC gains over \tgt{} ($\pm$ standard deviation of difference) on duplicate question detection task on various target topics. AAEME is the auto-encoder meta-embedding of \citet{BaoB2018}.}
\end{table}

\begin{table}[ht]
\setlength\tabcolsep{3.0pt}
\resizebox{\hsize}{!}{
\begin{tabular}{|l|l|l|l|c|}
\hline
       & \multicolumn{3}{c|}{Ohsumed} 
        & \multicolumn{1}{c|}{20NG Avg}
       \\
Method & Micro & Macro & Rare & 5 topics \\
\hline
\tgt & 26.3 & 14.7 & 3.0 & 88.9 \\ \hline
\src & -1.0$_{\pm \text{0.9}}$ & 0.$_{\pm \text{0.5}}$ & 0.$_{\pm \text{0.1}}$ & -3.9$_{\pm \text{1.2}}$ \\
AAEME & -1.0$_{\pm \text{0.9}}$ & 0.$_{\pm \text{0.5}}$ & 0.$_{\pm \text{0.1}}$ & -3.9$_{\pm \text{1.2}}$ \\
\srcinit & 1.7$_{\pm \text{1.0}}$ & 1.8$_{\pm \text{1.7}}$ & 1.5$_{\pm \text{2.0}}$ & 0.0$_{\pm \text{1.6}}$ \\
\yangC & 0.6$_{\pm \text{0.5}}$ & 1.8$_{\pm \text{2.3}}$ & 3.7$_{\pm \text{4.7}}$ & - \\
\ourReg & 1.4$_{\pm \text{0.5}}$ & 2.5$_{\pm \text{1.2}}$ & 4.0$_{\pm \text{1.8}}$ & 0.4$_{\pm \text{1.3}}$ \\
\our & 2.0$_{\pm \text{0.9}}$ & 2.6$_{\pm \text{1.5}}$ & 1.1$_{\pm \text{1.4}}$ & 0.5$_{\pm \text{1.5}}$ \\
\rowcolor{green!10}
\our & \hl{2.3$_{\pm \text{0.7}}$} & \hl{3.4$_{\pm \text{1.3}}$} & \hl{4.3$_{\pm \text{1.2}}$} &  \hl{0.5$_{\pm \text{1.5}}$} \\ 
\rowcolor{green!10} +\ourReg & & & & \\ 
\hline

\end{tabular} 

}

\caption{Average accuracy gains over \tgt{} ($\pm$ std-dev) on Ohsumed and 20NG datasets.  We show macro and rare class accuracy gains for Ohsumed because of its class population skew.  Per-topic 20NG gains are in \tablename~\ref{tab:classify:ng_detailed} in Appendix.}
\label{tab:classify} 
\end{table}

\noindent
We next compare \srcinit, \yangC, and \ourReg{} on the three tasks: perplexity in \tablename~\ref{tab:perplexity}, duplicate detection in \tablename~\ref{tab:auc}, and classification in \tablename~\ref{tab:classify}. All three methods are better than baselines \src\ and \concat, which are much worse than \tgt\ indicating the presence of significant concept drift.  \citet{regemnlp} provided no comparison between \yangC{} (their method) and \srcinit; we find the latter slightly better.
\noindent
On the supervised tasks, \yangC\ is often worse than \tgt\ provided \tgt\ is allowed to train for enough epochs.  If the same number of epochs are used to train the two methods, one can reach the misleading conclusion that \tgt\ is worse.  \ourReg\ is better than \srcinit\ and \yangC\ particularly with respect to perplexity, and rare class classification (Table~\ref{tab:classify}).  We conclude that a well-designed word stability-based regularizer can improve upon epoch-based fine-tuning.

\begin{table}[ht]
\centering
\setlength\tabcolsep{3.0pt}
\begin{tabular}{|l|l|l|l|l|}
\hline
 & Physics & Gaming & Android & Unix  \\
\hline
\multicolumn{5}{|c|}{\yangC's reduction in Perplexity over \tgt}   \\
\hline
Original & 1.1$_{\pm \text{1.1}}$ & 1.5$_{\pm \text{1.2}}$ & 0.9$_{\pm \text{0.1}}$ & 0.7$_{\pm \text{0.8}}$  \\
+SrcInit & \hl{2.1$_{\pm \text{0.9}}$} & \hl{5.7$_{\pm \text{0.8}}$} & \hl{1.1$_{\pm \text{0.5}}$} & \hl{2.1$_{\pm \text{0.8}}$}  \\
\hline
\multicolumn{5}{|c|}{\yangC's gain in AUC over \tgt}   \\
\hline
Original & -1.2$_{\pm \text{0.4}}$ & 0.1$_{\pm \text{0.1}}$ & \hl{-0.2$_{\pm \text{0.1}}$} & \hl{-0.4$_{\pm \text{0.1}}$} 
\\ 
+SrcInit & \hl{-0.4$_{\pm \text{0.2}}$} & \hl{2.4$_{\pm \text{0.2}}$} & -0.5$_{\pm \text{0.5}}$ & -0.5$_{\pm \text{0.2}}$ 
\\
\hline
\end{tabular}
\caption{Effect of initializing with source embeddings.  We show mean gains over \tgt\ over 9 runs ($\pm$ std-dev).}
\label{tab:init}
\end{table}

\paragraph*{Impact of source initialization}
\tablename~\ref{tab:init} compares \tgt\ and \yangC\ with two initializers: (1)~random  as proposed by \citet{regemnlp}, and (2)~with source embeddings.  \yangC\ after source initialization is better in almost all cases.  \our\ and \ourReg\ also improve with source initialization, but to a smaller extent.  (More detailed numbers are in \tablename~\ref{tab:init:all} of Appendix.)  We conclude that initializing with pretrained embeddings is helpful even with regularizers.

\paragraph*{Comparison with Meta-embeddings} In Tables~\ref{tab:auc} and ~\ref{tab:classify} we show results with the most recent meta-embedding method AAEME.
AAEME provides gains over \tgt\ in only two out of six cases\footnote{On the topic classification datasets in Table~\ref{tab:classify}, AAEME and its variant DAEME were worse than \src.  We used the dev set to select the better of \src\ and their best method.}.

\subsection{Performance of \our}

We next focus on the performance of \our\ on all three tasks: perplexity in \tablename~\ref{tab:perplexity}, duplicate detection in \tablename~\ref{tab:auc}, and classification in \tablename~\ref{tab:classify}.  \our\ is always among the best two methods for perplexity.  In supervised tasks, \our\ is the only method that provides significant gains for all topics: AUC for duplicate detection increases by 2.4\%, and classification accuracy increases by 1.4\% on average. \our+\ourReg\ performs even better than \our\ on all three tasks particularly on rare words. An ablation study on other variants of \our\ appear in the Appendix.
\begin{table}[ht]
\centering
\setlength\tabcolsep{3.0pt}
\begin{tabular}{l|l|l|l|l|l}
  Pair                &     Tgt & Src & Reg & Reg  &   Src \\ 
                      &         & Tune& Freq & Sense & Sel\\\hline
Unix topic &  & & & & \\ \hline
nice, kill & 4.6 & 4.5 & 4.4 & 4.4 & 5.2 \\
vim, emacs & 5.7 & 5.8 & 5.7 & 5.8 & 6.4 \\                      
print, cat & 5.0 & 4.9 & 4.9 & 5.0 & 5.4 \\
kill, job & 5.2 & 5.1 & 5.2 & 5.3 & 5.8 \\
make, install & 5.1 & 5.1 & 5.3 & 5.7 & 5.8 \\
character, unicode & 4.9 & 5.1 & 4.7 & 4.6 & 5.8 \\ \hline

Physics topic &  & & & & \\ \hline
lie, group         & 5.2 & 5.0 & 4.4 & 5.1 & 5.8 \\
current, electron  & 5.3 & 5.3 & 4.7 & 5.3 & 5.7 \\
potential, kinetic & 5.8 & 5.8 & 4.5 & 5.9 & 6.1 \\
rotated, spinning & 5.0 & 5.7 & 6.0 & 5.1 & 5.6 \\
x-ray, x-rays & 5.3 & 7.0 & 6.1 & 5.5 & 6.4 \\
require, cost & 4.9 & 6.2 & 5.2 & 5.1 & 5.3 \\
cool, cooling & 5.6 & 6.0 & 6.4 & 5.7 & 5.7 \\
\hline
\end{tabular}
\caption{\label{tab:anecdotes} Example word pairs and their normalized similarity across different methods of training embeddings.}
\end{table}

\paragraph{Word-pair similarity improvements:} In \tablename~\ref{tab:anecdotes}, we show  normalized\footnote{We sample a set $S$ of 20 words based on their frequency. Normalized similarity between $a$ and $b$ is $\frac{\cos(a,b)}{\sum_{w \in (S \cup b)}\cos(a,w)}$. Set $S$ is fixed across methods.} cosine similarity of word pairs pertaining to the Physics and Unix topics.  Observe how word pairs like ({\em nice, kill}), ({\em vim, emacs})
in Unix and 
({\em current, electron}), ({\em lie, group})
in Physics are brought closer together as a result of importing the larger unix/physics subset from~$\DS$.  In each of these pairs,  words (e.g. {\em nice, vim, lie, current}) have a different prominent sense in the source (Wikipedia).  Hence, methods like \srcinit, and \ourReg\ cannot help.  In contrast, word pairs like (cost, require), (x-ray, x-rays) whose sense is the same in the two corpus benefit significantly from the source across all methods.

\paragraph{Running time:} \our\ is five times slower than \yangC, which is still eminently practical. $\widehat{\DS}$ was within $3\times$ the size of $\DT$ in all domains.  If $\DS$ is available, \our\ is a practical and significantly more accurate option than adapting pretrained source embeddings. \our+\ourReg\ complements \our\ on rare words, improves perplexity, and is never worse than \our.

\begin{table}[ht]
\setlength\tabcolsep{3.0pt}
\resizebox{\hsize}{!}{
  \begin{tabular}{|l|r|r|r|r|r|}
    \hline
    & Physic & Game & Andrd & Unix & Med(Rare) \\
    \hline
    \tgt & 89.7 & 88.4 & 89.4 & 89.2 & 9.4 \\
    \hline
    \srcinit & $-0.2$ & 0.6 & $-0.4$ & $-0.2$
    & $-2.1$ \\
   \our  & $1.9$ & 0.5 & 0.0 & $-0.2$
    & 1.1 \\
    \hline
  \end{tabular}
  }
  \caption{
  \label{tab:large:c}Performance with a larger target corpus size of 10MB on the four deduplication tasks (AUC score) and one classification task (Accuracy on rare class). More details in Table~\ref{tab:large} of Appendix.}
\end{table}

\paragraph*{Effect of target corpus size}
The problem of importing source embeddings is motivated only when target data is limited. When we increase target corpus 6-fold, the gains of \our\ and \srcinit\ over \tgt\ was insignificant in most cases. However, infrequent classes continued to benefit from the source as shown in~\tablename~\ref{tab:large:c}.

\begin{table}[th]
\setlength\tabcolsep{3.0pt}
\resizebox{\hsize}{!}{
  \begin{tabular}{|l|r|r|r|r|r|}
    \hline
    & Physic & Game & Andrd & Unix & Med \\
    \hline
    \tgt & 86.7 & 82.6 & 86.8 & 85.4 & 26.3 \\
    \hline
    ELMo & $-1.0$ & \hl{4.5} & $-1.5$ & $-2.3$
    & 3.2 \\
    +\tgt & $-0.8$ & 3.8 & 0.5 & 0.0 
    & 4.1 \\
    +\srcinit & \cellcolor{white} $-0.5$ & 3.0 & 0.3 & 0.2
    & 3.5 \\
    +\our & \hl{2.6} & 4.1 & \hl{1.1} & \hl{1.5} 
    & \hl{4.6} \\
    \hline
  \end{tabular}
  }
  \caption{\label{tab:ctxt}Gains over \tgt\ with contextual embeddings on duplicate detection (columns 2--5) and classification (column 6).  (Std-dev in \tablename~\ref{tab:ctxt:std} of Appendix.)}
\end{table}

\subsection{Contextual embeddings}

We explore if contextual word embeddings obviate the need for adapting source embeddings, in the ELMo \citep{PetersNIGCKZ2018ELMo} setting, a contextualized word representation model, pre-trained on a 5.5B token corpus\footnote{\url{https://allennlp.org/elmo}}. 
We compare ELMo's contextual embeddings as-is, and also after concatenating them with each of \tgt, \srcinit, and \our\ embeddings in Table~\ref{tab:ctxt}.  First, ELMo+\tgt\ is better than \tgt\ and ELMo individually.  This shows that contextual embeddings are useful but they do not eliminate the need for topic-sensitive embeddings.  Second, ELMo+\our\ is better than ELMo+\tgt.  Although \our\ is trained on data that is a strict subset of ELMo, it is still instrumental in giving gains since that subset is aligned better with the target sense of words.
We conclude that topic-adapted embeddings can be useful, even with ELMo-style contextual embeddings.


Recently, BERT~\cite{devlin2018bert} has garnered a lot of interest for beating contemporary contextual embeddings on all the GLUE tasks. 
We evaluate BERT on question duplicate question detection task on the four StackExchange topics. We use pre-trained BERT-base, a smaller 12-layer transformer network, for our experiments. We train a classification layer on the final pooled representation of the sentence pair given by BERT to obtain the binary label of whether they are duplicates. This is unlike the earlier setup where we used EMD on the fixed embeddings.

To evaluate the utility of a 
relevant topic focused corpus, we fine-tune the pre-trained checkpoint either on $\DT$ (\srcinit) or on $\DT \cup \widehat{\DS}$ (\ourR) using BERT's masked language model loss. The classifier is then initialized with the fine-tuned checkpoint. Since fine-tuning is sensitive to the number of update steps, we tune the number of training steps using performance on a held-out dev set.  F1 scores corresponding to different initializing checkpoints are shown in table~\ref{tab:bert_ft}. It is clear that  pre-training the contextual embeddings on relevant target corpus helps in the downstream classification task. However, the gains of \ourR\ over \tgt\ is not clear. This could be due to incomplete or noisy sentences in $\widehat{\DS}$. There is need for more experimentation and research to understand the limited gains of \ourR\ over \srcinit\ in the case of BERT. We leave this for future work.


\begin{table}
\setlength\tabcolsep{3.0pt}
\centering
\begin{tabular}{|l|r|r|r|r|}
\hline
Method & Physics & Gaming & Android & Unix \\
\hline
BERT & 87.5 & 85.3 & 87.4 & 82.7 \\
\srcinit & {\bf 88.0} & {\bf 89.2} & 88.5 & 83.5 \\
\ourR & 87.9 & 88.4 & {\bf 88.6} & {\bf 85.1} \\
\hline
\end{tabular}
\caption{F1 scores on question de-duplication task using BERT-base and when fine-tuned on Tgt only ($\DT$) and Tgt and selected source ($\DT \cup \widehat{\DS}$)}
\label{tab:bert_ft}
\end{table}
\section{Conclusion}
\label{sec:End}

We introduced one regularization and one source-selection method for adapting word embeddings from a partly useful source corpus to a target topic.  They work better than recent embedding transfer methods, and give benefits even with contextual embeddings.  It may be of interest to extend these techniques to embed knowledge graph elements.

\medskip \noindent
\paragraph{Acknowledgment:} Partly supported by an
IBM AI~Horizon grant. We thank all the anonymous reviewers for their constructive feedback.

\bibliographystyle{acl_natbib}
\bibliography{ML,voila}

\newpage
\twocolumn[\section*{\centering \papertitle \\ (Appendix)}]

Here we present additional results that did not fit into the main paper.

\begin{table}[htb]
\centering
\begin{tabular}{|l|r|r|r|} \hline
 & Tokens & Vocab size & \# duplicates\\
\hline
Physics & 542K & 6,026 & 1981\\
Gaming & 302K & 6,748 & 3386\\
Android & 235K & 4,004 & 3190\\
Unix & 262K  & 6,358& 5312\\
\hline
\end{tabular}
\caption{\label{tab:stats}Statistics of the Stack Exchange data used in duplicate question detection.  $\DS$ has a vocabulary of 300,000 distinct words.}
\end{table}

\begin{table}[htb]
\centering
\setlength\tabcolsep{0.7pt}
\begin{tabular}{|l|l|l|l|l|l|}
\hline 
Method & Physic & Gamng & Andrd & Unix & Med \\
\hline
\tgt & 121.9 & 185.0 & 142.7 & 159.5 & 158.9 \\
\hline
\srcinit & $2.3_{\pm 0.7}$&$6.8_{\pm 0.3}$&$1.1_{\pm 0.3}$&$3.1_{\pm 0.0}$&$5.5_{\pm 0.8}$\\
\yangC &$2.1_{\pm 0.8}$&$7.1_{\pm 0.7}$&$1.8_{\pm 0.4}$&$3.4_{\pm 0.5}$&$6.8_{\pm 0.9}$\\
RegSens &$5.0_{\pm 0.1}$&$13.8_{\pm 0.3}$&$6.7_{\pm 0.8}$&$9.7_{\pm 0.3}$&$14.6_{\pm 1.0}$\\
\our &$5.8_{\pm 0.9}$&$11.7_{\pm 0.6}$&$5.9_{\pm 1.2}$&$6.4_{\pm 0.1}$&$8.6_{\pm 3.0}$\\
\hline
\our + &$6.2_{\pm 1.3}$&$12.5_{\pm 0.3}$&$7.9_{\pm 1.8}$&$9.3_{\pm 0.2}$&$10.5_{\pm 0.9}$\\
RegSens & & & & & \\
\hline
\end{tabular}
\caption{\label{tab:perplexity_with_sd}Average reduction in language model perplexity over \tgt on five StackExchange Topics. ($\pm$ standard deviation) are shown in supplementary.}
\end{table}

\begin{table}[!ht]
\setlength\tabcolsep{1.0pt}
\centering
  \begin{tabular}{|l|l|l|l|l|l|}
    \hline
    & Physic & Game & Andrd & Unix & Med \\
    \hline
    \tgt & 86.7 & 82.6 & 86.8 & 85.4 & 26.3 \\
    \hline
    Elmo & -1.0$_{\pm \text{0.4}}$ & 4.5$_{\pm \text{0.3}}$ & -1.5$_{\pm \text{0.8}}$ & -2.3$_{\pm \text{0.3}}$ 
    & 3.2$_{\pm \text{1.3}}$ \\
    +\tgt & -0.8$_{\pm \text{0.4}}$ & 3.8$_{\pm \text{0.4}}$ & 0.5$_{\pm \text{0.1}}$ & -0.0$_{\pm \text{0.1}}$ 
    & 4.1$_{\pm \text{1.5}}$ \\
    +ST & -0.5$_{\pm \text{0.3}}$ & 3.0$_{\pm \text{0.2}}$ & 0.3$_{\pm \text{0.5}}$ & 0.2$_{\pm \text{0.2}}$ 
    & 3.5$_{\pm \text{0.6}}$ \\
    +\our & 2.6$_{\pm \text{0.5}}$ & 4.1$_{\pm \text{0.1}}$ & 1.1$_{\pm \text{0.4}}$ & 1.5$_{\pm \text{0.2}}$ 
    & 4.6$_{\pm \text{0.9}}$ \\
    \hline
  \end{tabular}
  \caption{\label{tab:ctxt:std}AUC scores comparing contextual embeddings on the question dedup tasks and Medical Abstract Classification task. Shown in the first row is the performance of \tgt\ and the rows below show mean gains over \tgt\ ($\pm$ std-dev). ST refers to \srcinit\ .}
\end{table}

\paragraph{Ablation studies on \our}
We compare variants in the design of \our\ in Tables~\ref{tab:ablation:class}, \ref{tab:wect:ablation_lm} and ~\ref{tab:wect:ablation_auc}. In \ourR\ we run the SrcSel without weighting the source snippets by the $Q(w,C)$ score in \eqref{eq:SnippetSelect}.  We observe that the performance is worse than with the $Q(w,C)$ score. Next, we check if the score would suffice in down-weighting irrelevant snippets without help from our IR based selection.  In \ourC\ we include 5\% random snippets from $\DS$ in addition to those in \our\ and weigh them all by their $Q(w,C)$ score.  We find in Table~\ref{tab:wect:ablation_auc} that the performance drops compared to \our.  Thus, both the $Q(w,C)$ weighting and the IR selection are important components of our source selection method.

\begin{table}[htb]
\centering
\begin{tabular}{|l|l|l|} \hline
 & Micro Accuracy & Macro Accuracy \\
\hline
\tgt & 26.3$_{\pm 0.5}$ & 14.7$_{\pm 1.2}$ \\
\hline
\ourR & 27.3$_{\pm 0.3}$ & 16.1$_{\pm 1.6}$ \\
\our & 28.3$_{\pm 0.4}$ & 17.3$_{\pm 0.7}$ \\ \hline
\end{tabular}
\caption{\label{tab:ablation:class}Ablation studies on \our\ for classification on the Medical dataset. }
\end{table}

\begin{table}[ht]
\setlength\tabcolsep{1.0pt}
\centering
\begin{tabular}{|l|l|l|l|l|} \hline
 & \multicolumn{4}{|c|}{LM Perplexity}  \\ \hline
 & Physics & Gaming & Android & Unix \\
\hline
\tgt & 121.9$_{\pm 0.6}$ & 185.0$_{\pm 0.3}$ & 142.7$_{\pm 2.7}$ & 159.5$_{\pm 1.2}$ \\ \hline
SSR & 114.8$_{\pm 0.2}$ & 172.7$_{\pm 1.5}$ & 131.6$_{\pm 0.7}$ & 151.8$_{\pm 1.1}$ \\
\our & 116.1$_{\pm 0.9}$ & 173.3$_{\pm 0.6}$ & 136.7$_{\pm 1.1}$ & 153.1$_{\pm 0.1}$ \\
\hline
\end{tabular}
\caption{\label{tab:wect:ablation_lm} Ablation studies on \our\ over LM perplexity. SSR refers to \ourR\ .}
\end{table}

\begin{table}[ht]
\setlength\tabcolsep{2pt}
\centering
\begin{tabular}{|l|l|l|l|l|} \hline
 & \multicolumn{4}{|c|}{Question Dedup: AUC} \\ \hline
 & Physics & Gaming & Android & Unix \\
\hline
\tgt & 86.7$_{\pm 0.4}$ & 82.6$_{\pm 0.4}$ & 86.8$_{\pm 0.5}$ & 85.3$_{\pm 0.3}$ \\ \hline
\ourR & 89.2$_{\pm 0.2}$ & 85.6$_{\pm 0.4}$ & 87.5$_{\pm 0.3}$ & 86.8$_{\pm 0.2}$ \\
\ourC & 88.7$_{\pm 0.3}$ & 84.8$_{\pm 0.3}$ & 87.0$_{\pm 0.5}$ & 85.8$_{\pm 0.3}$ \\
\our & 90.4$_{\pm 0.2}$ & 85.4$_{\pm 0.5}$ & 87.4$_{\pm 0.4}$ & 87.5$_{\pm 0.1}$ \\
\hline
\end{tabular}
\caption{\label{tab:wect:ablation_auc} Ablation studies on \our\ over AUC.}
\end{table}

\begin{table}
\centering
\setlength\tabcolsep{1.0pt}
\begin{tabular}{|l|l|l|l|l|l|}
\hline
 &
Sci & Com & Pol & Rel & Rec 
\\ \hline
\tgt & 92.2 & 79.9 & 94.8 & 87.3 & 90.3  \\\hline
\src
 & -0.1$_{\pm \text{0.8}}$ & -9.1$_{\pm \text{0.7}}$ & -3.3$_{\pm \text{0.9}}$ & -1.0$_{\pm \text{2.3}}$ & -6.0$_{\pm \text{0.5}}$ 
\\
ST & 0.0$_{\pm \text{1.1}}$ & 0.0$_{\pm \text{1.3}}$ & -0.1$_{\pm \text{1.5}}$ & 0.1$_{\pm \text{2.5}}$ & 0.2$_{\pm \text{1.1}}$  \\
RS & 0.9$_{\pm \text{0.9}}$ & -0.2$_{\pm \text{1.3}}$ & 0.2$_{\pm \text{1.3}}$ & 1.2$_{\pm \text{2.0}}$ & 0.1$_{\pm \text{0.8}}$  \\
\our & 1.2$_{\pm \text{0.8}}$ & 0.1$_{\pm \text{1.4}}$ & 0.5$_{\pm \text{1.2}}$ & 0.5$_{\pm \text{2.6}}$ & 0.3$_{\pm \text{1.0}}$  \\\hline
\end{tabular}
\caption{\label{tab:classify:ng_detailed} Average accuracy gains over \tgt\ and $\pm$std-dev on five classification domains: Science, Computer, Politics, Religion, Recreation, from 20 NG dataset. ST and RS abbreviate to \srcinit\ and \ourReg\ respectively.}
\end{table}

\paragraph{Critical hyper-parameters}
The number of neighbours $K$ used for computing embedding based stability score as shown in~(\ref{eqn:sim_score}) is set to 10 for all the tasks. We train each of the different embedding methods for a range of different epochs: \{5, 20, 80, 160, 200, 250\}. The $\lambda$ parameter of \ourReg\ and \yangC\ is tuned over \{0.1, 1, 10, 50\}. Pre-trained embeddings $\ES$ are obtained by training a CBOW model for 5 epochs on a cleaned version of 20160901 dump of Wikipedia. All the embedding sizes are set to $300$. 


\onecolumn
\begin{table}
\centering
\setlength\tabcolsep{3.0pt}
\begin{tabular}{|l|r|r|r|r|r|r|r|r|} \hline
  & \multicolumn{4}{|c|}{Perplexity} & \multicolumn{4}{|c|}{AUC}  \\
\hline
 & Physics & Gaming & Android & Unix & Physics & Gaming & Android & Unix \\
\hline
\tgt & 121.9  & 185.0 & 142.7 & 159.5 & 86.7 & 82.6 & 86.8 & 85.3 \\
\hline
\yangC & 2.1$_{\pm \text{0.8}}$ & 7.0$_{\pm \text{0.7}}$ & 1.8$_{\pm \text{0.4}}$ & 3.4$_{\pm \text{1.0}}$ & -0.4$_{\pm \text{0.4}}$ & 2.3$_{\pm \text{0.3}}$ & -0.6$_{\pm \text{0.4}}$ & -0.3$_{\pm \text{0.3}}$ \\
\yangC-rinit & -1.6$_{\pm \text{0.9}}$ & 1.2$_{\pm \text{0.6}}$ & 1.6$_{\pm \text{0.2}}$ & 2.6$_{\pm \text{0.8}}$ & -1.2$_{\pm \text{0.2}}$ & 0.$_{\pm \text{0.3}}$ & -0.2$_{\pm \text{0.5}}$ & -0.3$_{\pm \text{0.3}}$ \\
\hline
\ourReg & 5.0$_{\pm \text{0.1}}$ & 13.8$_{\pm \text{0.3}}$ & 6.7$_{\pm \text{0.7}}$ & 9.7$_{\pm \text{0.8}}$ & -0.3$_{\pm \text{0.3}}$ & 2.1$_{\pm \text{0.4}}$ & -0.6$_{\pm \text{0.3}}$ & -0.3$_{\pm \text{0.5}}$ \\
\ourReg-rinit & 3.6$_{\pm \text{1.0}}$ & 11.1$_{\pm \text{0.5}}$ & 7.0$_{\pm \text{1.2}}$ & 8.9$_{\pm \text{1.4}}$ & 0.7$_{\pm \text{0.2}}$ & 1.2$_{\pm \text{0.2}}$ & -0.3$_{\pm \text{0.6}}$ & -0.2$_{\pm \text{0.5}}$ \\
\hline
\our & 5.8$_{\pm \text{0.9}}$ & 11.7$_{\pm \text{1.8}}$ & 6.0$_{\pm \text{1.2}}$ & 6.3$_{\pm \text{1.0}}$ & 3.7$_{\pm \text{0.2}}$ & 2.8$_{\pm \text{0.5}}$ & 0.6$_{\pm \text{0.4}}$ & 2.2$_{\pm \text{0.1}}$ \\
\ourR-rinit & 5.8$_{\pm \text{1.0}}$ & 12.5$_{\pm \text{0.4}}$ & 10.4$_{\pm \text{1.4}}$ & 7.9$_{\pm \text{1.0}}$ & 2.5$_{\pm \text{0.1}}$ & 2.$_{\pm \text{0.5}}$ & 0.4$_{\pm \text{0.4}}$ & 1.5$_{\pm \text{0.2}}$ \\
\hline
\end{tabular}
\caption{\label{tab:init:all} Source initialization vs random initialization (with suffix {\it -rinit}) on \yangC\ , \ourReg\ , \our\ . Shown in the table is the average gain over \tgt\ for each method.}

\begin{tabular}{|l|l|l|l|l|l|l|}
\hline
Method & Physics & Gaming & Android & Unix & Med (Micro) & Med (Rare) \\
\hline
\tgt & 89.7$_{\pm \text{0.3}}$ & 88.4$_{\pm \text{0.2}}$ & 89.4$_{\pm \text{0.3}}$ & 89.2$_{\pm \text{0.3}}$ & 31.4$_{\pm \text{0.9}}$ & 9.4$_{\pm \text{3.0}}$ \\
\srcinit & 89.5$_{\pm \text{0.2}}$ & 89.0$_{\pm \text{0.2}}$ & 89.0$_{\pm \text{0.2}}$ & 89.0$_{\pm \text{0.2}}$ & 31.3$_{\pm \text{0.4}}$ & 7.3$_{\pm \text{2.6}}$ \\
\our & 91.6$_{\pm \text{0.3}}$ & 88.9$_{\pm \text{0.8}}$ & 89.4$_{\pm \text{0.2}}$ & 89.0$_{\pm \text{0.2}}$ & 31.3$_{\pm \text{0.9}}$ & 10.5$_{\pm \text{1.8}}$ \\
\hline
\end{tabular}
\caption{\label{tab:large}Performance with a larger target corpus size of 10MB on the four deduplication tasks (AUC score) and one classification task (Accuracy) shown in the last two columns.}
\end{table}

\end{document}